\documentclass[preprint,12pt]{elsarticle}

\usepackage{amssymb}
\usepackage{multirow}
\usepackage{amsmath}
\usepackage{url}
\usepackage{arydshln}
\journal{Nuclear Physics B}

\newcommand{\KPNoiseA}{\text{KP-Noise}_{1\text{--}4}^{20\%}}

\newcommand{\KPNoiseB}{\text{KP-Noise}_{5\text{--}9}^{20\%}}

\newcommand{\KPNoiseC}{\text{KP-Noise}_{1\text{--}4}^{50\%}}

\newcommand{\KPNoiseD}{\text{KP-Noise}_{5\text{--}9}^{50\%}}

\begin{document}

\begin{frontmatter}



\title{The Blind Spot in 2D Infants' Pose Estimation:
Robust Learning from Noisy Annotations}
% \titlerunning{The Blind Spot in 2D Infants' Pose Estimation}


\author[1]{Emanuele Cardinale\corref{cor1}}
\ead{emanuele.cardinale@phd.unich.it}
\cortext[cor1]{Corresponding author}
\author[2]{Marco Proietti}
\author[3]{Alessandro Cacciatore}
\author[4]{Maria Francesca Spadea}
\author[5]{Lucia Migliorelli\fnref{fn1}}
\author[6]{Sara Moccia\fnref{fn1}}

\affiliation[1]{organization={Department of Engineering and Geology, Università degli Studi ``G. d’Annunzio'' Chieti-Pescara},
                city={Pescara},
                postcode={65127},
                country={Italy}}

\affiliation[2]{organization={Department of Political Science, Communication and International Relations, University of Macerata},
                city={Macerata},
                postcode={62100},
                country={Italy}}

\affiliation[3]{organization={Laboratory of Computational Oncology, Department of Oncology, KU Leuven},
                city={Leuven},
                postcode={3000},
                country={Belgium}}

\affiliation[4]{organization={Institute of Biomedical Engineering, Karlsruhe Institute of Technology},
                city={Karlsruhe},
                postcode={76131},
                country={Germany}}

\affiliation[5]{organization={Department of Political Science, Università di Teramo},
                city={Teramo},
                postcode={64100},
                country={Italy}}

\affiliation[6]{organization={Department of Innovative Technologies in Medicine and Dentistry, Università degli Studi ``G. d’Annunzio'' Chieti-Pescara},
                city={Chieti},
                postcode={66100},
                country={Italy}}

\fntext[fn1]{Lucia Migliorelli and Sara Moccia share senior authorship.}

\begin{abstract}
Noisy annotations pose a significant challenge for supervised deep learning, as neural networks rely on large-scale, high-quality labeled data whose corruption can severely impair model performance. Although robustness to label noise has been extensively studied for classification tasks, it remains relatively underexplored in Pose Estimation (PE).
This limitation becomes critical in clinical contexts, including neonatology, where PE of preterm infants is used to support the assessment of spontaneous motility, a key indicator of neurodevelopmental trajectories. In such settings, infants' images labeling is further hindered by visual challenges (e.g., keypoint self-occlusions, caregiver interference), making the annotation process inherently susceptible to errors. To tackle noisy annotations in PE, we introduce REliable keypoint selection via Memory of traINing Dynamics (REMIND), a clustering-based keypoint-selection strategy that exploits keypoint-wise training dynamics to identify noisy labels without assuming any prior knowledge of the noise distribution, thus enabling noise-free model training.
When evaluated on the proprietary NeoPose dataset, comprising 46 videos of 46 preterm infants recorded in real clinical settings, REMIND correctly identifies noisy annotations across multiple corruption scenarios, achieving up to 93\% Area Under the Curve (AUC) with three different PE architectures used in the relevant literature. To our knowledge, this is the first study to explicitly address label noise in preterm infants' PE, paving the way for the design of trustworthy learning-based algorithms for infants'monitoring support when data quality cannot be guaranteed.

\end{abstract}

% \begin{highlights}

% \item REMIND: an unsupervised approach for detecting noisy labels in neonatal pose estimation through keypoint-wise training dynamics.
% \item Validated on a proprietary dataset of 46 preterm infants recorded in routine clinical settings.
% \item Achieved up to 93\% AUC in noisy-label identification across multiple corruption scenarios.
% \item First study addressing label noise in preterm infants' pose estimation for trustworthy neonatal monitoring.
% \end{highlights}


\end{frontmatter}

\section{Introduction}
% Survivors of preterm birth are at increased risk of neurodevelopmental impairment and disability.1 Five to 10% of infants born preterm before the 32nd week of gestation suffer from major neurologic disorders, including cerebral palsy (CP) and severe intellectual disability.2 Early identification of such babies remains a challenge but is important, considering that there is a potential benefit of early intervention when the brain is most responsive to repair due to plasticity.3, 4, 5, 6

The quality of spontaneous movements in preterm infants during the first months of life provides critical insights into the development of their nervous system \cite{gm_neurosviluppo1,gm_neurosviluppo2}. Deviations from these endogenously generated activities, known as general movements, serve as reliable early markers for Neurodevelopmental Disorders (NDDs), including cerebral palsy and Autism Spectrum Disorder (ASD) \cite{gma_autism1,gma_autism2,gma_autism3}. This is particularly important in preterm infants, who are at a significantly higher risk of neurodevelopmental impairment and long-term disability \cite{preterm_gma}. Today the gold standard for evaluating these movements is the visual observation by trained clinical experts \cite{gma_autism4}. Despite its high predictive validity, clinical implementation is hampered by several bottlenecks - the method is subjective, requires extensive expert-based training and is time-consuming to perform.  Early automated efforts relied on wearable sensors, which, however, had the downfall to be be intrusive, potentially alter spontaneous movement patterns and cause discomfort, particularly in medically fragile preterm infants \cite{gma_sensor1,gma_sensor2}.

% The quality of spontaneous infant movements during the first months of life provides critical insight into the development of the nervous system \cite{gm_neurosviluppo1,gm_neurosviluppo2}. These endogenously generated activities, known as general movements (GMs), follow a distinct developmental timeline, transitioning from large-body "writhing" movements in early infancy to small, irregular "fidgety" movements between two and five months of age. Deviations from these patterns serve as reliable early markers for neurodevelopmental disorders (NDDs), including cerebral palsy and autism spectrum disorder (ASD) \cite{gma_autism1,gma_autism2,gma_autism3}. The gold standard for evaluating these movements is the General Movement Assessment (GMA), a clinical tool based on the visual perception of trained experts \cite{gma_autism4}. Despite its high predictive validity, clinical implementation is hampered by several bottlenecks - the method is subjective, requires extensive expert-based training and is time-consuming to perform.  Early automated efforts relied on wearable sensors, which ,however, have the downfall to be be intrusive, potentially alter spontaneous movement patterns and cause discomfort \cite{gma_sensor1,gma_sensor2}.
% %

\begin{figure}[tbp!]
     \centering
     \includegraphics[width=\linewidth]{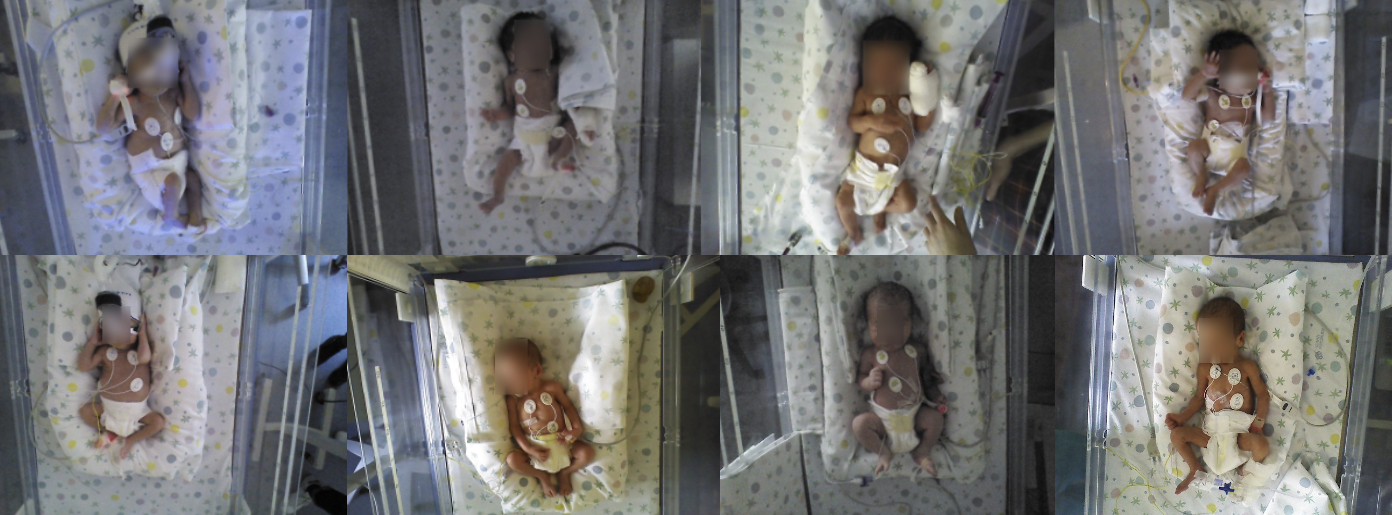}
     \caption{Sample images from the NeoPose dataset showing the presence of self-occlusions, medical devices, casts, and caregivers’ or practitioners’ hands.}
     \label{fig:neopose}
\end{figure}

In recent years, vision-based approaches for infant pose estimation (PE) have gained increasing attention as unobtrusive tools to support clinicians in the assessment of spontaneous movements and current research is highlighting the need to adapt state-of-the-art, adult-centric PE models to infant-specific data \cite{comparison2,finetuning2,finetuning3}. Most approaches in this direction rely on fully supervised training, where the quality of annotations plays a crucial role in determining model performance. Yet, annotation quality is often compromised by the fact that the process itself is repetitive and inherently prone to error ~\cite{annotation_pose_quality}. This observation aligns with preliminary studies that report the presence of annotation noise in widely used human PE datasets and highlights the detrimental influence of noisy labels (i.e., keypoints annotated at incorrect anatomical locations) on PE-model performance \cite{influencefaultylabels}. In neonatal PE, the annotation process is further challenged by strong visual ambiguities arising from self-occlusions, medical devices, casts, and the presence of caregivers’ or practitioners’ hands, which can hinder accurate keypoint localization (Fig.~\ref{fig:neopose}). While learning with noisy labels has become a well-established research area in deep learning, limited attention has been reserved to learning under noisy annotations in PE, compared to other domain (e.g., image classification), where a broad spectrum of noisy-label learning strategies has been developed, including specialised loss functions and sample selection methods ~\cite{song2025survey}. Many of these approaches are grounded in the Small-Loss (SL) hypothesis \cite{coteaching,small_loss_2}, which assumes that samples incurring lower training loss during the early stages of training are more likely to be correctly labeled, following the memorization effect of deep networks \cite{memorization_effect}. To the best of our knowledge, the only work addressing noisy labels in keypoint regression proposes ScarceNet \cite{scarcenet}, a framework for animal pose pseudo-annotation that primarily relies on the small-loss trick, which discards samples with high instantaneous regression loss based on a predefined threshold. This threshold must be tuned according to the assumed noise level in the dataset, limiting its practicality. Moreover, recent studies \cite{lstm_noisylabels,latestopping} have shown that instantaneous loss is an unreliable proxy for label quality, as it exhibits high variability across training epochs, suggesting that the temporal evolution of the loss — referred to as training dynamics— provides more stable and informative signals for distinguishing clean from noisy annotations.

\begin{figure}[tbp]
     \centering
     \includegraphics[width=.95\linewidth]{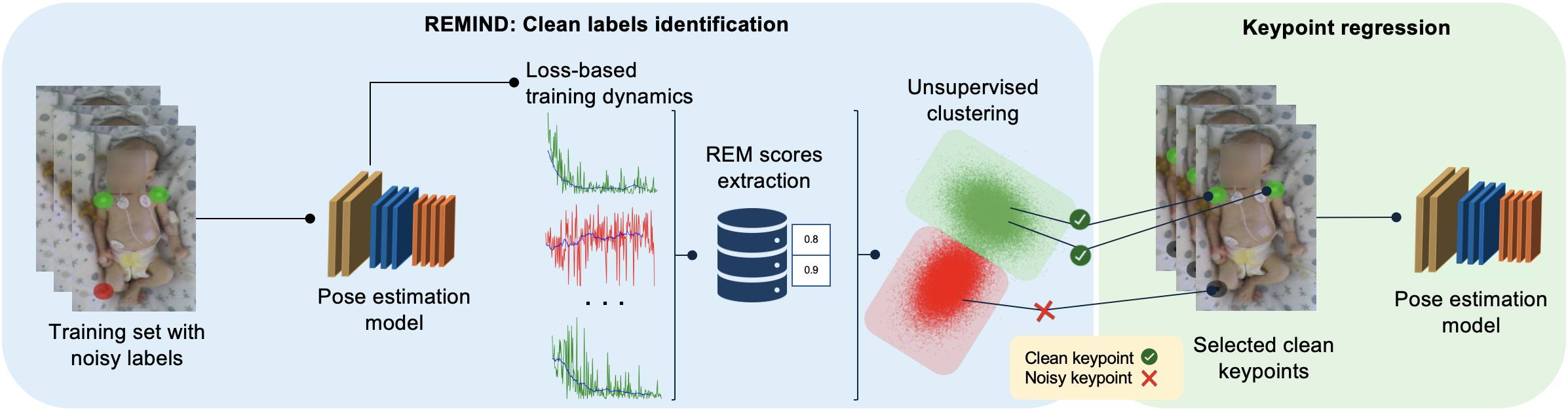}
     \caption{Proposed REMIND strategy. Clean keypoints are identified via clustering using keypoint-wise features ($REM$ scores) extracted from training-loss dynamics.}
     \label{fig:workflow}
\end{figure}

The main contributions of the work are summarised hereafter:

\begin{itemize}
\item We introduce REMIND (REliable keypoint selection via Memory of traINing Dynamics -- Fig.~\ref{fig:workflow}), a novel unsupervised strategy that leverages the temporal evolution of keypoint-wise loss values to identify noisy annotations. Unlike SL-based approaches, REMIND does not rely on predefined filtering thresholds. \item We validate REMIND on NeoPose, a newly collected dataset of preterm infant videos acquired during routine clinical practice, currently among the largest datasets specifically focused on hospitalized preterm infants.
\end{itemize}

The rest of the paper is organised as follows: Section \ref{sec:rw} reviews the state of the art on infant PE and noisy labels, Section \ref{sec:meth} introduces the REMIND method and describes the experimental protocol adopted for its evaluation, Section \ref{sec:res} presents the results and discussion, Section \ref{sec:conc} concludes the paper.

% Based on this consideration, we here introduce REMIND (REliable keypoint selection via Memory of traINing Dynamics - Fig. \ref{fig:workflow}), a novel unsupervised strategy that leverages the temporal evolution of keypoint-wise loss values to identify noisy annotations. Unlike SL-based approaches, REMIND does not rely on predefined filtering thresholds or prior assumptions about noise distribution. We validate REMIND on NeoPose, a newly collected dataset of preterm infant videos acquired during routine clinical practice, currently among the largest datasets specifically focused on hospitalized preterm infants. riformula elenco puntato.

\section{Related work}
\label{sec:rw}

\subsection{Infants Pose Estimation}

 % Branching from the broader field of human pose estimation, infant pose estimation aims to recover the spatial configuration of an infant’s body by localizing a predefined set of anatomical keypoints from an image.Unlike general human pose estimation, infant pose estimation presents additional challenges due to differences in anatomical proportions, smaller body scale, and the constrained and atypical nature of infant postures. A growing body of research \cite{comparison1,comparison2,comparison3} has focused on evaluating and benchmarking generic human pose estimation models originally trained on adult datasets, including and not limited to OpenPose \cite{openpose}, HRNet \cite{hrnet}, AlphaPose \cite{alphapose}, and ViTPose \cite{vitpose}, often reporting the latter as one of the best-performing general-purpose models for infants in supine positions. Nevertheless, the performance of these methods typically degrades when applied to infants because of the aforementioned domain differences \cite{comparison2}. To bridge this gap, researchers have explored various fine-tuning and domain adaptation strategies inspired by recent advances in deep learning. 

 A growing body of research \cite{comparison1,comparison2,comparison3} has focused on evaluating and benchmarking generic human PE models originally trained on adult datasets on infants' images, including and not limited to OpenPose \cite{openpose}, HRNet \cite{hrnet}, AlphaPose \cite{alphapose}, and ViTPose \cite{vitpose}, often reporting the latter as one of the best-performing general-purpose models for infants in supine positions. Nevertheless, the performance of these methods typically degrades when applied to infants because of several domain differences \cite{comparison2}. Branching from the broader field of human PE, a primary challenge in adapting adult models to infants lies in the substantial anatomical and distributional shifts between these two populations. Infants has different body proportions compared to adults, characterized by shorter limbs and a much larger head-to-torso ratio. Furthermore, infants exhibit unique pose distributions (such as bending legs over the chest) that are rarely captured in large-scale adult datasets like COCO \cite{coco} or MPII \cite{mpii}.

 To bridge this gap, researchers have explored various fine-tuning and domain adaptation strategies inspired by recent advances in deep learning. Jahn et al. \cite{comparison2} fine-tune high-performing adult backbones such as HRNet, OpenPose, and ViTPose on a relatively small, specialized dataset comprising approximately 4,500 annotated frames from 75 recordings of 75 infants aged 4 to 16 weeks, under different recording setups. Cao et al. \cite{aggpose} introduce AggPose, a framework that extends multi-scale transformer-based architectures to infant PE, and adopt a 21-keypoint representation defined by clinicians to capture finer-grained movements beyond the standard 17 COCO keypoints. They use for training a proprietary dataset made of 20748 labeled images. Huang et al. \cite{fidip} propose the Fine-tuned Domain-adapted Infant Pose framework alongside the SyRIP dataset (1700 annotated images), which combines real and synthetic infant poses, with synthetic data generated using the SMIL model \cite{modello_3d_bambini}. Their approach leverages a domain confusion network to transfer knowledge from adult datasets while aligning feature representations between synthetic and real domains. Grafton et al. \cite{alex_grafton_neonatal} shift the attention to the complexities of infant PE in neonatal intensive care units, where occlusions from blankets, medical equipment, and ongoing interventions are frequent. They fine-tune COCO-pretrained PE models on a dataset of 24 infants and demonstrate that multimodal signal fusion—combining RGB, depth, and infrared inputs at early, intermediate, and late stages of the network—can improve robustness in these complex settings. Recognizing that labeled data are often scarce in clinical contexts due to privacy constraints and the labor-intensive nature of annotation, Bose et al. \cite{finetuning2} investigate unsupervised domain adaptation methods and introduce SHIFT, a framework that adopts a pseudo-labeling-based mean teacher strategy to adapt adult pose estimators to infant data. To better account for domain-specific challenges, it incorporates an infant manifold pose prior that penalizes physically implausible configurations.

% For instance, \cite{openpose_finetuning} retrained OpenPose on a dataset of annotated infant images, achieving a 60\% reduction in mean error. Similarly, \cite{alex_grafton_neonatal} employed transfer learning by adapting models pretrained on adult datasets such as MS COCO to a limited neonatal clinical dataset, adding information clues coming from depth and IR images. thereby improving pose estimation accuracy despite the scarcity of labeled infant data. In \cite{fidip}, the authors introduce the Fine-tuned Domain-adapted Infant Pose (FiDIP) framework together with the SyRIP dataset, which includes both real and synthetic infant images. The synthetic samples are generated using the SMIL model \cite{modello_3d_bambini}, motivated by the difficulty of collecting large-scale infant datasets due to privacy and legal constraints. Likewise, \cite{finetuning3} presents a framework that combines pretrained backbones, such as HRNet, with synthetic infant data to align feature distributions between synthetic and real domains, leading to improved performance in low-label settings. AggPose \cite{aggpose} adopts a transformer-based architecture with deep feature aggregation and follows a pretraining-and-fine-tuning paradigm in which knowledge learned from large-scale adult datasets is adapted to infant-specific characteristics. More recently, SHIFT \cite{finetuning2} introduced an unsupervised domain adaptation framework that leverages synthetic adult data and infant manifold priors to generate more plausible infant pose predictions.

\subsection{Learning with noisy labels}
It is now well established that deep neural networks and noisy labels are uneasy partners, as the high capacity of over-parameterized models allows them to eventually memorize incorrect annotations, leading to a substantial degradation in generalization performance on unseen data \cite{zhang2017_noisylabels}. However, seminal studies have shown that deep neural networks do not memorize data indiscriminately \cite{memorization_effect}. Instead, during the early stages of training, they tend to prioritize learning simple and generalizable patterns from clean samples, a phenomenon commonly referred to as the memorization effect, before progressively fitting noisy examples in later epochs. Building on this observation, the work in \cite{coteaching} introduced the SL hypothesis, which posits that samples consistently exhibiting low training loss are more likely to be correctly labeled. This insight has motivated a wide range of approaches for noisy-label learning, particularly methods based on sample selection strategies that exploit early-epoch losses or, more generally, the temporal evolution of loss during training, often referred to as training dynamics. Jia et al. \cite{lstm_noisylabels} use an LSTM-based detector to automatically identify mislabeled samples by analyzing raw sequences of training dynamics rather than manually designed features. Yuan et al. \cite{latestopping} introduce the First-time k-epoch Learning metric to detect noise based on how many epochs a sample requires before being consistently classified correctly. Wang et al. \cite{chronoselect} propose ChronoSelect, which uses a four-stage temporal memory and trajectory analysis to partition data into clean, boundary, and noisy subsets. This line of research has predominantly focused on classification tasks, with relatively few studies addressing regression problems \cite{noisy_labels_regression1,noisy_labels_regression2,noisy_labels_regression3}. Further narrowing the scope to human PE, the field remains largely underexplored, with only a few notable works such as~\cite{influencefaultylabels} and \cite{scarcenet}. The former provides empirical evidence on the impact of annotation quality on human PE performance, while the latter is the closest to our approach, explicitly tackling the problem of learning from noisy pseudo-labels in animal PE by incorporating a threshold-based, small-loss sample selection strategy in the first stage of its pipeline. Building upon this line of research, our work moves beyond threshold-based filtering by leveraging the training dynamics of individual keypoints. Specifically, we exploit the temporal evolution of keypoint-wise loss values to perform unsupervised clustering and identify potentially noisy annotations, without requiring predefined filtering thresholds.

\section{Methods}
\label{sec:meth}

\subsection{REMIND}
%paragraph{\textbf{REMIND.}}
\label{subsec:remind}

% Let us consider a dataset \(\mathcal{D}\) for HPE. The goal is training a neural model $f$ by minimizing a heatmap-based loss, i.e., the Mean Squared Error ($MSE$).
Let us consider a neural model ($f$) for PE and a dataset $D\{(I_j,P_j)\}_{j=1}^{N}$, where $I_j \in \mathbb{R}^{H \times W \times 3}$ denotes an RGB image of size $H\times W$, $P_j \in \mathbb{R}^{H \times W \times K}$ the corresponding ground-truth joint heatmaps set, $K$ the number of annotated keypoints, and $N$ the number of images in $D$. For each sample $j$, Gaussian heatmaps $P_j$ derive from 2D ground-truth joints coordinates $\{(x_{j,k}, y_{j,k})\}_{k=1}^{K}$, to provide a more robust supervision signal than direct coordinate regression.
% to make learning more robust than direct coordinate regression. 
%

To model annotation noise, we define a dataset $\tilde{D}\{(I_j, \tilde{P_j})\}_{j=1}^{N}$ corrupted by noise, where $\tilde{P_j}$ denotes a potentially corrupted version of $P_j$. $\tilde{D}$ may contain a mix of clean and noisy labels for each sample, reflecting real annotation conditions. 

Firstly, $f$ is trained on $\tilde{D}$ by minimizing a heatmap-based loss, i.e., the Mean Squared Error ($MSE$) between $\tilde{P_j}$ and $\hat{P_j} = f(I_j)$. The $MSE$ values $l_{j,k}^{(e)}$ are recorded at each epoch $e$ for each keypoint $k$ in each sample $j$, forming the loss trajectory vector $\mathbf{L}_{j,k} = [l_{j,k}^{(0)}, l_{j,k}^{(1)}, \dots, l_{j,k}^{(E-1)}] \in \mathbb{R}^E$, where E is the number of training epochs. To reduce stochastic fluctuations, $\mathbf{L}_{j,k}$ is smoothed using a moving average filter. From our experimental observations, clean keypoints typically exhibit a consistent decrease in loss over training and tend to reach their minimum loss at later stages. In contrast, noisy keypoints often show limited loss reduction and more irregular learning trajectories. Based on these observations, we define the REMIND score ($REM$) for the $k$\textsubscript{th} keypoint of the $j$\textsubscript{th} sample:
% , and represents the keypoint-wise behaviour of the model throughout the learning process over training epochs ($E$).
% for the \( k \)-th keypoint in the \(i\)-th sample, 

\begin{equation}
    {REM}_{j,k} = \Delta l_{j,k} + \Delta t_{j,k}
\end{equation} 
where $\Delta l_{j,k}$ captures how much the model learns from the keypoint, measured by the normalized peak-to-trough loss value drop:

\begin{equation}
    \Delta l_{j,k} = \frac{\max(\mathbf{L}_{j,k}) - \min(\mathbf{L}_{j,k})}{\max(\mathbf{L}_{j,k}) + \min(\mathbf{L}_{j,k})}
\end{equation} 
and $\Delta t_{j,k}$ captures temporal information on when the minimum keypoint-wise loss is reached: 

\begin{equation}
   \Delta t_{j,k} = \frac{\operatorname{arg\,\underset{e}{\min}} \mathbf{L}_{j,k} - \operatorname{arg\,\underset{e}{\max}} \mathbf{L}_{j,k}}{E}
\end{equation}

After obtaining the $REM$ score for each $k$\textsubscript{th} keypoint of each $j$\textsubscript{th} sample, we train K-means to cluster keypoints in the 2D space defined by $(\Delta t$, $\Delta l)$. Since, empirically, both $\Delta t$ and $\Delta l$ tend to decrease as annotation noise increases, the cluster characterised by the lower centroid is identified as containing noisy keypoints. After clustering, $f$ is trained on a de-noised version of $\tilde{D}$, in which keypoints assigned to the noisy cluster are excluded by setting their COCO visibility flag to zero, so that they do not contribute to the loss computation. 
% the selected clean keypoints. The 
It is worth noting that unlike work on sample selection in classification (e.g., \cite{xia2021sample}) our method operates at the keypoint level and does not discard entire samples, even when they partially contain noisy annotations.

\subsection{Noise taxonomy and injection}

\begin{figure}[tbp!]
     \centering
     \includegraphics[width=0.8\linewidth]{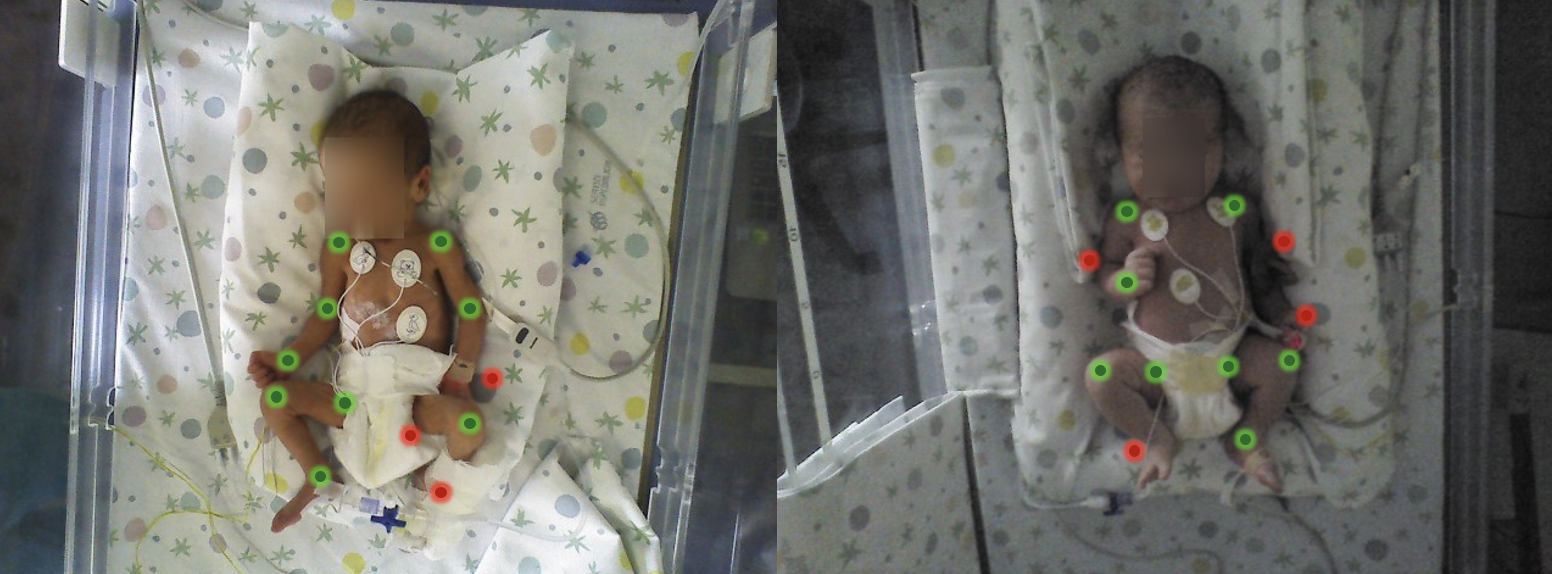}
     \caption{Examples of correct (green) and Gaussian-perturbed keypoints (red).}
     \label{fig:example_noise}
\end{figure}

%\paragraph{\textbf{Noise taxonomy and injection.}}
\label{subsec:noise}
To obtain $\tilde{D}$, in our experiments we randomly select a subset of $N'$ samples from $D$ and randomly choose a subset of keypoints $K_j$ for each selected sample. We artificially inject noise into ground-truth 2D joint coordinates $(x_{\tilde{j},\tilde{k}}, y_{\tilde{j},\tilde{k}})$ with $\tilde{j} = 1,\dots,N'$ and $\tilde{k} = 1,\dots,K_j$ by adding Gaussian perturbations $\epsilon_x, \epsilon_y \sim \mathcal{N}(0, \sigma^2)$:

\begin{equation}
\tilde{x}_{\tilde{j},\tilde{k}} = x_{\tilde{j},\tilde{k}} + \epsilon_x, \qquad
\tilde{y}_{\tilde{j},\tilde{k}}= y_{\tilde{j},\tilde{k}} + \epsilon_y.
\end{equation}

Figure \ref{fig:example_noise} illustrates an example image with both noisy and clean keypoint annotations superimposed. This noise scenario, modeled as local shifts induced by Gaussian perturbations, follows prior work on this topic \cite{noisy_pose_estimation2011}. We experimentally define four noise configurations by combining two image-level corruption rates (20\% or 50\% of the training samples) with two keypoint-level perturbation ranges (1–4 or 5–9 corrupted keypoints per image): $\vphantom{\KPNoiseA}\KPNoiseA$, $\vphantom{\KPNoiseB}\KPNoiseB$, $\vphantom{\KPNoiseC}\KPNoiseC$, and $\vphantom{\KPNoiseD}\KPNoiseD$. The percentage of noisy keypoints in the training set ranges from a minimum of 3.01\% ($\vphantom{\KPNoiseA}\KPNoiseA$) to a maximum of 20.57\% ($\vphantom{\KPNoiseD}\KPNoiseD$).

\subsection{Dataset}
%\paragraph{\textbf{Experiments and dataset.}}
In this work, we rely on the NeoPose dataset\footnote{The dataset is available upon request, subject to approval and legal constraints.}, a collection of 5456 RGB frames from 46 RGB-D videos of preterm infants spontaneously moving, recorded right before discharge in the neonatology department of the G. Salesi Hospital in Ancona, Italy. Infants exhibit variability in gestational age ($31.87 \pm 3.77$), weight ($2 \pm 0.79$ kg), and length ($44.13 \pm 4.12$ cm). Videos are recorded using an Astra Mini S-Orbbec RGB-D camera, positioned approximately $\sim$50 cm above the crib, with a spatial resolution of 640×480 pixels and a frame rate of 30 fps. Each video refers to a single infant, recorded shortly before hospital discharge. Manual annotation was carried out by a team of neonatologists following the standard COCO keypoint annotation format. For obtaining $\tilde{D}$, we set $\sigma$ to 10\% of the diagonal of the bounding box enclosing the infant in the specific image.

\subsection{Metrics}
To evaluate the performance of REMIND in detecting noisy annotations, we use standard classification metrics, including Area Under the ROC Curve ($AUC$), Sensitivity ($Sens$), Specificity ($Spec$), and Precision ($Prec$).  We further evaluate REMIND consistency across models using an Agreement Index (AI) defined as the proportion of noise-corrupted keypoints that are correctly identified as noisy by all three models for each noise configuratioN. Since REMIND is based on unsupervised clustering, ground-truth labels are used only for evaluation purposes.
The Euclidean distance of each keypoint from the clean-cluster centroid in the $(\Delta t, \Delta l)$ space is interpreted as a confidence score:
\begin{equation}
d_{j,k} = \sqrt{(\Delta t_{j,k} - \mu_{\Delta t})^2 + (\Delta l_{j,k} - \mu_{\Delta l})^2}
\end{equation}

By thresholding this score, we compute the $AUC$ with respect to ground-truth noise labels. To evaluate the performance of PE models trained on the fully clean, noise-corrupted, and REMIND de-noised datasets,we use Mean Average Precision ($mAP$), i.e., $AP$ calculated at different Object Keypoint Similarity thresholds from 0.50 to 0.95 in steps of 0.05, formally:
\begin{equation}
mAP = \frac{1}{T} \sum_{t=1}^{T} AP_{OKS_t}, \quad OKS_t \in \{0.50, 0.55, \dots, 0.95\}
\end{equation}

% To evaluate the performance of REMIND in detecting noisy annotations, we use Area Under the ROC Curve ($AUC$), Sensitivity ($Sens$), Specificity ($Spec$), Precision ($Prec$). We further evaluate REMIND consistency across models with an agreement index defined as the proportion of noise-corrupted keypoints that were correctly identified as noisy by all three models for each noise configuration. Since REMIND is based on unsupervised clustering, ground-truth labels are used only for evaluation. The Euclidean distance of each keypoint from the clean-cluster centroid in the $(\Delta t$, $\Delta l)$ space is interpreted as a confidence score. By thresholding this score, we compute the $AUC$ with respect to ground-truth noise labels. To evaluate the perfomance of PE models trained on the fully-clean, noise-corrupted and REMIND de-noised dataset, we use Mean Average Precision ($mAP$), i.e., $AP$ calculated at different Object Keypoint Similarity thresholds from 0.50 to 0.95 in steps of 0.05.

\subsection{Experiments}
We conduct our experiments on HRNet, ResNet 101 \cite{resnet}, and ViTPose, 3 models widely used in infants' PE \cite{comparison1,comparison2}. All models are trained for 210 epochs on a NVIDIA Ampere A100 GPU (64GB) using the OpenMMLab Pose Framework\footnote{\url{https://github.com/open-mmlab/mmpose}}. ViTPose uses AdamW as optimizer and layer-wise learning rate decay (12 layers, decay rate 0.75) with gradient clipping (max norm 1.0), while HRNet and ResNet 101 are optimized with Adam, following standards in the literature. All models use a batch size of 64 and we apply a linear warm-up during the first 500 iterations, followed by a multi-step learning rate schedule with reductions at epochs 170 and 200\footnote{The code will be made public upon publication.}.

REMIND is compared against the state-of-the-art SL trick. Following the protocol proposed in \cite{scarcenet}, 
SL is activated after an initial warm-up phase, allowing the model to partially converge before sample selection. Once activated, SL selects a predefined proportion of keypoints with the highest instantaneous loss values and designates them as noisy, excluding them from subsequent training. 

\subsubsection{Cross-Domain Dataset Validation}

% Se la mettessimo come sottosezione in risultati e discussioni? nope deve essere anticipato nel protocollo sperimentale non si può arrivare con effetto sorpresa 

% ok ho riadattato un po' la parte che prima era nei risultati, ora modifico un po' i risultati.

The proposed REMIND primarily relies on the exploitation of keypoint-wise training dynamics, therefore demonstrating  potential to generalize effectively across different keypoint regression scenarios. In this regard, we evaluate it on additional datasets drawn from domains beyond infant PE. The first dataset is a subset of SurgPose \cite{surgpose}, which consists of surgical scene recordings annotated with 14 keypoints corresponding to surgical instruments. We extract 5,000 frames to ensure a dataset size comparable to our infant PE dataset. The second dataset involves a different imaging modality and consists of lateral cephalogram radiographs annotated by clinical experts with 29 cephalometric landmarks \cite{cephalo}. Since these experiments are intended solely as a cross-dataset validation, we investigate a single keypoints regressor architecture, namely HRNet.

% the SL strategy is implemented by selecting an early starting training epoch and labeling a predefined fraction of keypoints with the highest loss values as noisy. 

\section{Results and discussion}

\label{sec:res}

\begin{table}[tbp!]
\centering
\caption{Noisy-keypoints identification performance ($AUC$, reported as percentages) across models and noise configurations using REMIND against small-loss trick (SL) filtering strategy~\cite{scarcenet}.}

\label{tab:kpnoise_results}
%{\fontsize{7}{8}\selectfont
\setlength{\tabcolsep}{6pt}
\renewcommand{\arraystretch}{1.15}
\resizebox{\textwidth}{!}{%
\begin{tabular}{c | c c c c c}
\hline
 &  & $\vphantom{\KPNoiseA}\KPNoiseA$  & $\vphantom{\KPNoiseB}\KPNoiseB$  & $\vphantom{\KPNoiseC}\KPNoiseC$  & $\vphantom{\KPNoiseD}\KPNoiseD$  \\
\hline
\multirow{2}{*}{ResNet 101}  
                         & SL     & 67.7 & 72.0 & 67.1 & 69.9 \\& REMIND & \textbf{97.1} & \textbf{96.7} & \textbf{95.3} & \textbf{93.3} \\
\hline
\multirow{2}{*}{HRNet}   
                         & SL     & 66.7 & 70.9 & 66.3 & 69.2 \\ & REMIND & \textbf{97.3} & \textbf{96.7} & \textbf{95.5} & \textbf{93.1} \\
\hline
\multirow{2}{*}{ViTPose} 
                         & SL     & 66.4 & 69.6 & 65.5 & 67.6 \\ & REMIND & \textbf{97.7} & \textbf{96.8} & \textbf{95.4} & \textbf{93.7} \\
\hline
\end{tabular}
}
%}
\end{table}

Table \ref{tab:kpnoise_results} summarises the results relevant to noisy-keypoint identification. Across all noise configurations and tested PE models, REMIND consistently outperforms the SL trick, with $AUC$ values ranging from 93.3\% to 97.7\% and coherently improving as the percentage of noisy keypoints decreases (i.e., from $\KPNoiseD$ to $\KPNoiseA$). Interestingly, the results show minimal variation across models, with $REM$ scores from ViTPose training emerging as the most informative to filter noisy keypoints. The agreement index, averaged across the four configurations, reaches 88.9\% for REMIND. This result indicates that REMIND provides substantially more stable noisy-keypoint identification across architectures and captures model-agnostic learning patterns associated with annotation noise, rather than architecture-specific artifacts.

% \begin{table}[tbp!]

\begin{figure}[tbp!]
     \centering
     \includegraphics[width=\linewidth]{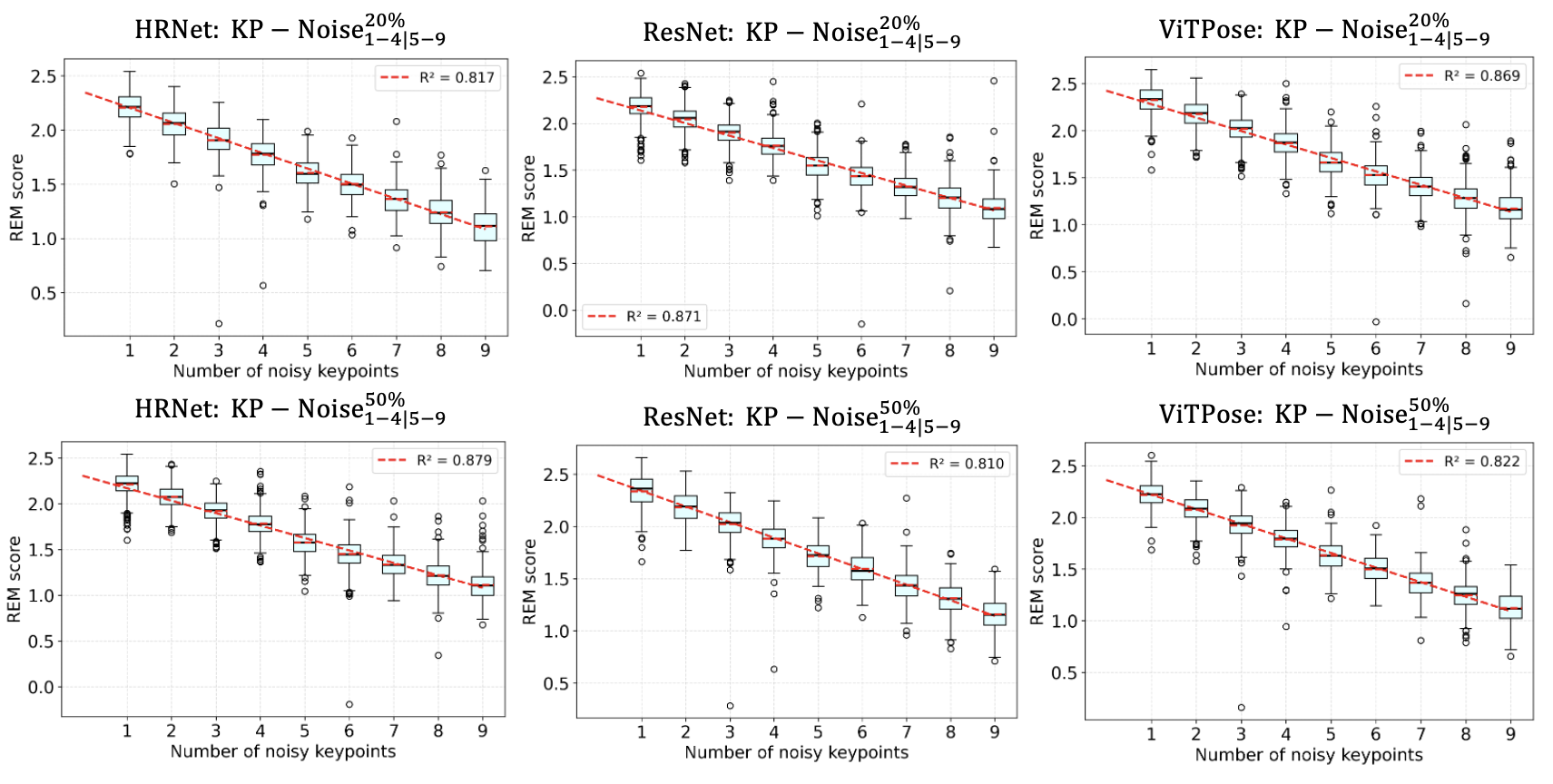}
     \caption{$REM$ score distributions versus the number of noise-corrupted keypoints per sample, showing a strong linear correlation (high R²) between noise severity and REMIND.}
     \label{fig:boxplots}
\end{figure}

\begin{figure}[tbp!]
     \centering
     \includegraphics[width=0.7\linewidth]{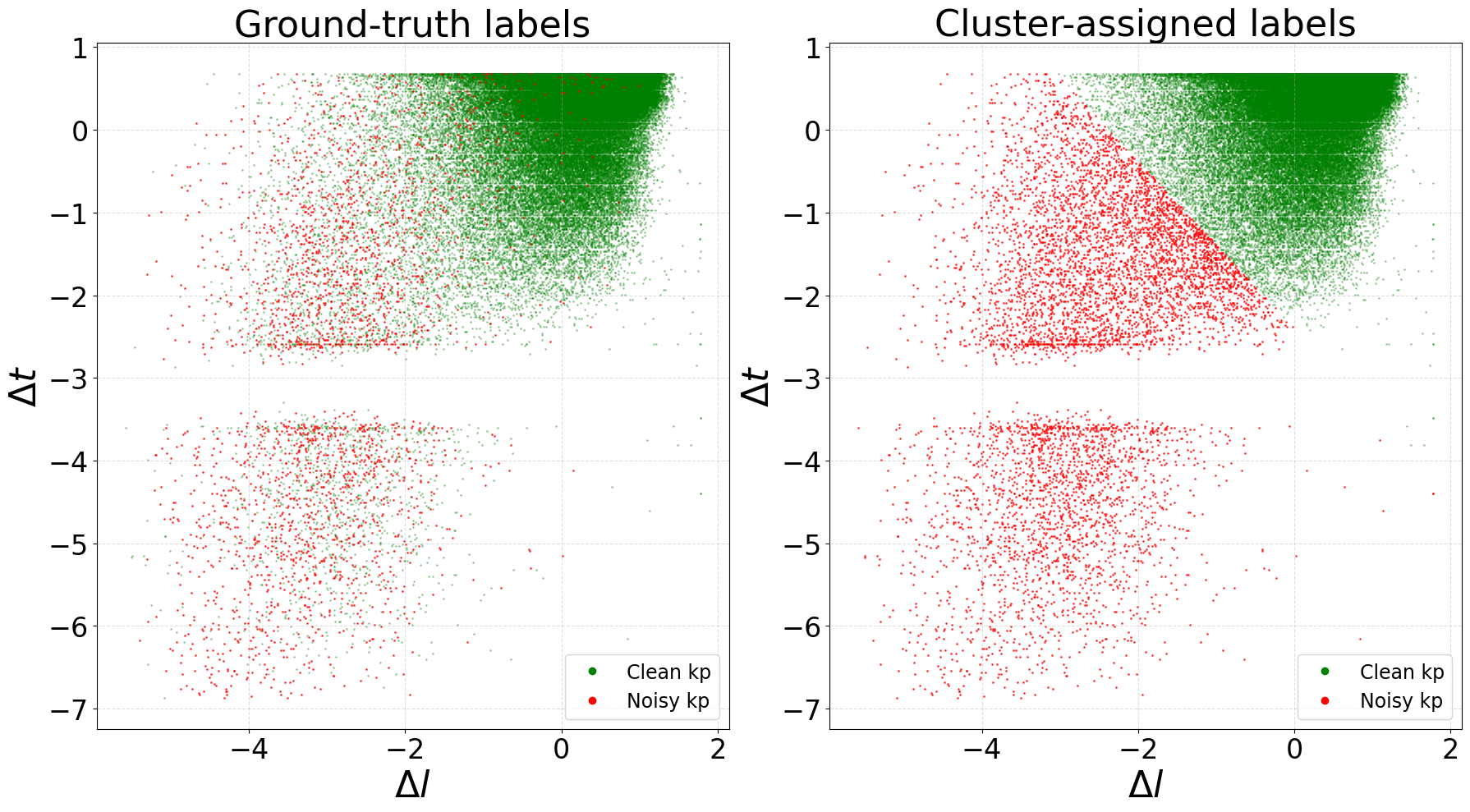}
     \caption{K-means clustering of $REM$ scores from HRNet trained under $\vphantom{\KPNoiseA}\KPNoiseA$. Keypoints are projected onto $(\Delta t, \Delta l)$. Red/green denote noisy/clean. Left: ground truth; right: cluster assignments.}
     \label{fig:clustering}
\end{figure}

The relationship between noise severity and $REM$ scores is shown in Fig.~\ref{fig:boxplots}. For each model and noise injection ratio, samples are grouped by the number of noise-corrupted keypoints, and the average $REM$ score is plotted. A linear regression analysis between the number of noisy keypoints per sample and the corresponding mean $REM$ score reveals a high coefficient of determination (R² ranging from 81\% to 87.9\%). This result proves that the proposed $REM$ score captures sufficient information to distinguish noisy keypoints from clean ones.

%To further investigate REMIND’s discriminative power, 

Fig. \ref{fig:clustering} shows the clustering results for HRNet under the $\KPNoiseA$ noise configuration. After applying K-Means clustering, a clear separation between keypoint types emerges. Noisy keypoints are consistently grouped within a well-defined region characterized by high values of both $\Delta l$ and $\Delta t$, enabling a highly effective clustering.

\begin{table}[tbp!]
\centering
\caption{$Sens$, $Spec$ and $Prec$ of noisy keypoints detection, after K-means clustering.}
\label{tab:conf_matrix_remind_clustering}
{\fontsize{7}{8}\selectfont
\setlength{\tabcolsep}{5pt}
\renewcommand{\arraystretch}{1.2}

\resizebox{\textwidth}{!}{%
\begin{tabular}{c | c c c | c c c}
\hline
 & \multicolumn{3}{c |}{$\KPNoiseA$} & \multicolumn{3}{c}{$\KPNoiseB$} \\
\hline
Metrics & ResNet 101 & HRNet & ViTPose & ResNet 101 & HRNet & ViTPose \\
\hline
$Sens$ & 90.4 & 91.1 & 87.7 & 88.9 & 88.8 & 87.0 \\
\hline
$Spec$ & 95.9 & 95.2 & 96.8 & 94.6 & 94.5 & 95.1 \\
\hline
$Prec$ & 40.8 & 37.1 & 46.3 & 59.9 & 59.5 & 61.6 \\
\hline\hline
& \multicolumn{3}{c |}{$\KPNoiseA$} & \multicolumn{3}{c}{$\KPNoiseB$} \\
\hline
Metrics & ResNet 101 & HRNet & ViTPose & ResNet 101 & HRNet & ViTPose \\
\hline
$Sens$ & 88.4 & 88.1 & 86.7 & 85.6 & 85.0 & 85.0 \\
\hline
$Spec$ & 93.0 & 92.6 & 93.6 & 87.9 & 87.7 & 88.4 \\
\hline
$Prec$ & 49.7 & 48.4 & 51.8 & 64.8 & 64.3 & 65.5 \\
\hline
\end{tabular}%
}
}
\end{table}

Table~\ref{tab:conf_matrix_remind_clustering} complements the visual results with quantitative metrics (i.e., $Sens$, $Spec$ and $Prec$) across all the architectures and noise configurations. $Sens$ and $Spec$ decrease as the amount of injected noise increases, whereas $Prec$ exhibits the opposite trend. This can be attributed to the fact that dataset imbalance (noisy-to-clean-ratio) rises from 3.01\% in $\KPNoiseA$ to 20.57\% in $\KPNoiseD$. However, $Prec$ remains consistently lower than $Sens$ and $Spec$, primarily due to the high false-positive rate, i.e., clean keypoints in the noisy cluster. This limitation may stem from the adoption of a hard binary clustering strategy (clean vs. noisy keypoints) not accounting for borderline cases such as hard or ambiguous samples \cite{forouzesh2024differences}.

\begin{figure}[tbp!]
     \centering
     \includegraphics[width=\linewidth]{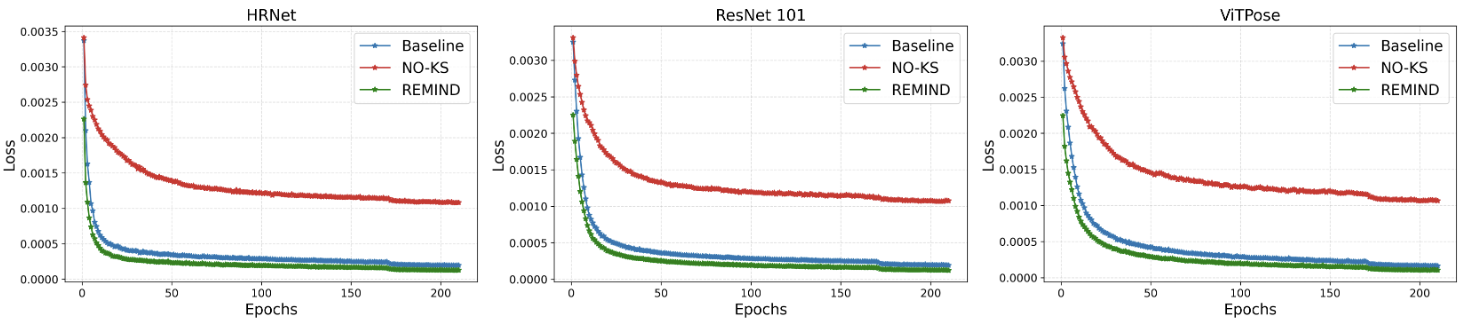}
     \caption{Training losses under three settings: (i) baseline, (ii) $\vphantom{\KPNoiseD}\KPNoiseD$ without keypoints selection (NO-KS), and (iii) $\vphantom{\KPNoiseD}\KPNoiseD$ with REMIND filtering.}
     \label{fig:training_loss}
\end{figure}

\begin{figure}[tbp!]
     \centering
     \includegraphics[width=1\linewidth]{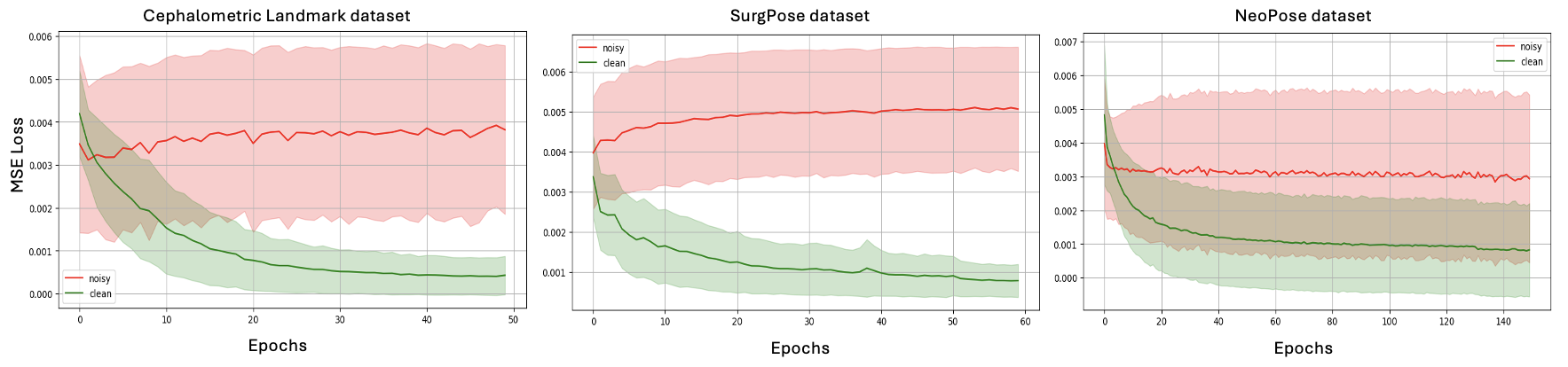}
     \caption{Averaged training losses trajectories for clean (green) and noisy (red) keypoints. Shaded bands indicate $\pm 1$ standard deviation.}
     \label{fig:loss_plot}
\end{figure}

\begin{table}[tbp!]
% \small
\centering
\caption{$mAP$ results. Baseline: training on the clean dataset; NO-KS: noisy training without keypoint selection; REMIND: proposed method. Bold indicates the best between NO-KS and REMIND.}
\label{tab:kpnoise_results_2}
{\fontsize{7}{8}\selectfont
\setlength{\tabcolsep}{5pt}
\renewcommand{\arraystretch}{1.2}

\resizebox{\textwidth}{!}{%
\begin{tabular}{c | c c c | c c c}
\hline
 & \multicolumn{3}{c |}{$\KPNoiseA$} & \multicolumn{3}{c}{$\KPNoiseB$} \\
\hline
 & ResNet 101 & HRNet & ViTPose & ResNet 101 & HRNet & ViTPose \\
\hline
Baseline     & 0.691 & 0.733 & 0.704 & 0.691 & 0.733 & 0.704 \\
\hdashline
NO-KS  & 0.612 & 0.525 & 0.641 & 0.584 & 0.598 & 0.617 \\
\hline
REMIND & \textbf{0.678} & \textbf{0.725} & \textbf{0.692}
       & \textbf{0.671} & \textbf{0.708} & \textbf{0.681} \\
\hline\hline
 & \multicolumn{3}{c |}{$\KPNoiseC$} & \multicolumn{3}{c}{$\KPNoiseD$} \\
\hline
 & ResNet 101 & HRNet & ViTPose & ResNet 101 & HRNet & ViTPose \\
\hline
Baseline     & 0.691 & 0.733 & 0.704 & 0.691 & 0.733 & 0.704 \\
\hdashline
NO-KS  & 0.536 & 0.581 & 0.603 & 0.431 & 0.397 & 0.382 \\
\hline
REMIND & \textbf{0.664} & \textbf{0.716} & \textbf{0.684}
       & \textbf{0.668} & \textbf{0.715} & \textbf{0.676} \\
\hline
\end{tabular}%
}
}
\end{table}

\begin{table}[tbp!]
\centering
\caption{$Sens$ and $Spec$ of noisy keypoints detection, after K-means clustering for SurgPose subset and the cephalometric landmarks dataset}
\label{tab:surgpose_cephalo}
{\fontsize{7}{8}\selectfont
\setlength{\tabcolsep}{5pt}
\renewcommand{\arraystretch}{1.2}

\resizebox{\textwidth}{!}{%
\begin{tabular}{c | c c | c c}
\hline
 & \multicolumn{2}{c |}{$\KPNoiseA$} & \multicolumn{2}{c}{$\KPNoiseB$} \\
\hline
Metrics & SurgPose & Cephalometric Landmark & SurgPose & Cephalometric Landmark \\
\hline
$Sens$  & 97.8 & 96.9 & 97.2 & 96.5 \\
\hline
$Spec$ & 98.3 & 99.4 & 98.9 & 98.8 \\

\hline\hline
& \multicolumn{2}{c |}{$\KPNoiseA$} & \multicolumn{2}{c}{$\KPNoiseB$} \\
\hline
Metrics & SurgPose & Cephalometric Landmark & SurgPose & Cephalometric Landmark \\
\hline
$Sens$ & 98.5 & 95.6 & 94.1 & 96.0 \\
\hline
$Spec$ & 99.1 & 98.2 & 98.6 & 96.5 \\

\hline
\end{tabular}%
}
}
\end{table}

We compare REMIND-filtered training with standard training on the fully noise-corrupted dataset without any keypoint selection. Fig. \ref{fig:training_loss} shows training losses for all three models under (i) the noise-free configuration, (ii) the hardest noisy configuration ($\KPNoiseD$), and (iii) the REMIND-filtered configuration. Training on noisy labels results in consistently higher loss values compared to the clean baseline, reflecting the detrimental effect of annotation noise on optimization. In contrast, applying REMIND substantially reduces this discrepancy, yielding a training loss trajectory that closely aligns with the clean-label baseline.

Table \ref{tab:kpnoise_results_2} reports $mAP$ values on the infant-level hold-out test set, preventing data leakage. Training with noisy samples has a detrimental impact on model performance, with an average $mAP$ drop across all configurations of $22.5\%$ $(\pm 10.7\%)$. In contrast, training after REMIND filtering limits this drop to only $1.8\%$ $(\pm 1.2\%)$, effectively preserving performance even under severe noise conditions. These results confirm the effectiveness of REMIND in mitigating the impact of annotation noise on PE performance.
%

% This detection framework is simple and interpretable, with the potential to generalize effectively across different keypoint regression scenarios. In this regard, we evaluate it on additional datasets drawn from domains beyond infant PE. The first dataset is a subset of SurgPose \cite{surgpose}, which consists of surgical scene recordings annotated with 14 keypoints corresponding to surgical instruments. We extract 5,000 frames to ensure a dataset size comparable to our infant PE dataset. The second dataset involves a different imaging modality and consists of lateral cephalogram radiographs annotated by clinical experts with 29 cephalometric landmarks \cite{cephalo}. 

Table \ref{tab:surgpose_cephalo} reports the quantitative results in terms of sensitivity and specificity for the cross validation datasets (SurgPose and cephalometric landmarks) under the four noise configurations. 

The results demonstrate that REMIND generalizes well across heterogeneous domains and highlight that performance on both the SurgPose subset and the cephalometric landmark dataset is consistently higher than that observed on infant PE. We attribute this difference primarily to the intrinsic complexity of the infant PE task. The NeoPose dataset is characterized by frequent self-occlusions, severe inter-part overlap, large pose variability, and ambiguous visual patterns caused by blankets, medical devices, or caregiver interactions. Moreover, Fig. \ref{fig:loss_plot} illustrates the training loss trajectories averaged over clean and noisy keypoints for a representative noise configuration ({$\KPNoiseA$}), showing a clear separation between the two groups during training for the SurgPose subset and the cephalometric landmark dataset. In comparison to the corresponding curves on the infant PE dataset, a noticeable reduction in separability is observed, further confirming the higher difficulty of the infant setting.

\section{Conclusion}
\label{sec:conc}

This work addresses the challenge of noisy annotations in infants' PE by introducing REMIND, an unsupervised sample selection strategy that leverages the evolution of the training dynamics to identify unreliable (and thus noisy) keypoint annotations. The dataset, although curated, remains limited in size and is currently collected from a single clinical center. To mitigate this limitation, we are going to start a new data collection protocol in collaboration with an additional clinical center to support multi-center validation.

Future work will investigate settings in which the validation set is also affected by annotation noise. This extension calls for the development of early stopping criteria that remain reliable even when validation annotations are partially corrupted~\cite{train_dyn1}. In order to assess the robustness and generalization of the proposed approach across different data distributions, we are currently working on expanding this work standard PE benchmarks such as COCO and MPII. A further relevant future direction concerns a more principled characterization of noise taxonomy. In this work, noise is currently assumed to follow a Gaussian shift around the true keypoint location, a formulation that is inherited from the work in \cite{noisy_pose_estimation2011} and remains a reasonable approximation. However, this assumption does not fully capture the structure of annotation noise in practice, where keypoints do not share the same probability of corruption. This issue is particularly evident in neonatal PE, which is strongly affected by visual challenges, and noisy annotations are often associated with inherently ambiguous cases, such as those arising from self-occlusion or visually uncertain keypoints. Future work will therefore focus on designing noise modeling strategies that explicitly account for such hard keypoints.

%\section{Declarations}

\subsection*{Acknowledgements}
We would like to acknowledge the ITALIAN FUND FOR APPLIED SCIENCES (FISA), grant number: FISA2022-00696, and ISCRA for awarding us access to the LEONARDO supercomputer, owned by the EuroHPC Joint Undertaking, hosted by CINECA (Italy).

\subsection*{Conflict of Interest}
The Authors has no known competitive interests. 

\subsection*{Ethics Approval and Consent to Participate}
The study received the approval of the Ethics Committee of the “Ospedali Riuniti di Ancona”, Italy (ID: Prot. 2019-399). Written informed consent was obtained from each infant's legal guardian. 

\subsection*{Data Availability}
The dataset used in this study will be made available upon request.

\bibliographystyle{splncs04}
\bibliography{biblio}

\end{document}